%% file: main.tex
\documentclass[10pt,twocolumn,letterpaper]{article}

\usepackage[pagenumbers]{cvpr}            %

\input{preamble}

\definecolor{cvprblue}{rgb}{0.21,0.49,0.74}
\usepackage[pagebackref,breaklinks,colorlinks,citecolor=cvprblue]{hyperref}

\def\paperID{2} %
\def\confName{CVPR}
\def\confYear{2025}

\title{Less Biased Noise Scale Estimation for Threshold-Robust RANSAC}

\author{Johan Edstedt$^1$
 \\
{\normalsize $^1$Linköping University}
}

\begin{document}
\twocolumn[{%
\centering
\renewcommand\twocolumn[1][]{#1}%
\maketitle
    \includegraphics[width=.96\linewidth]{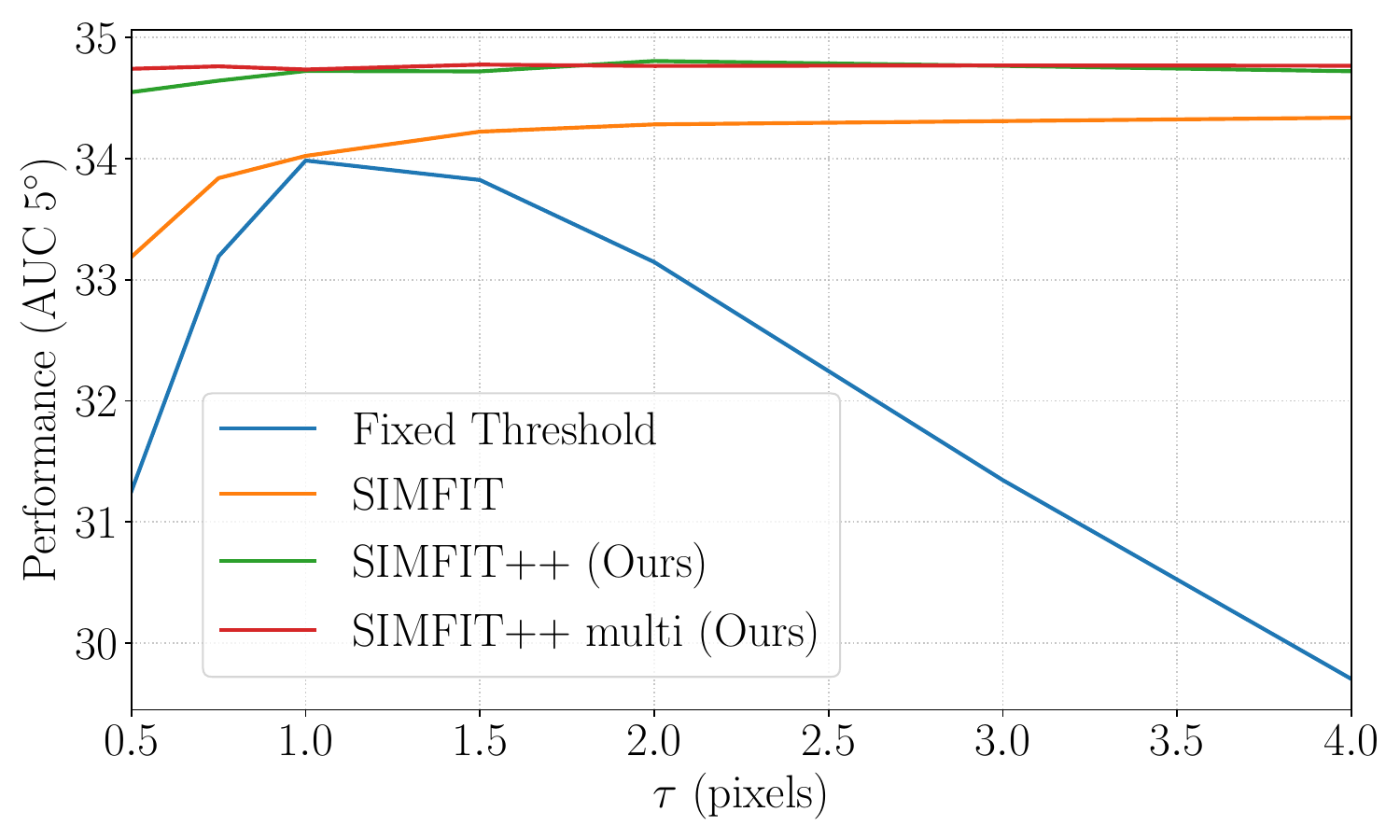}
    \captionof{figure}{\textbf{Fundamental matrix estimation performance as a function of RANSAC threshold (higher is better).} Traditionally, following an initial guess, RANSAC thresholds are not updated during optimization. In practice, this means that the threshold will be suboptimal, as different images and models require different thresholds. We overcome this issue by alternating between estimating the noise scale $\sigma$ and RANSAC, and using that $\tau = F^{-1}(\alpha) \sigma$ for the confidence level $\alpha \in [0,1]$, which we fix as $\alpha=0.99$.}
    \label{fig:main-results}}]
\maketitle
\input{sec/abstract}    
\input{sec/intro}

\input{sec/method}

\input{sec/results}

\input{sec/discussion}
\input{sec/conclusion}
\section*{Acknowledgements}
Thanks to Georg Bökman and Viktor Larsson for the discussions, and to the reviewers for the corrections and suggestions.
This work was supported by the Wallenberg Artificial
Intelligence, Autonomous Systems and Software Program
(WASP), funded by the Knut and Alice Wallenberg Foundation and by the strategic research environment ELLIIT, funded by the Swedish government. The computational resources were provided by the
National Academic Infrastructure for Supercomputing in
Sweden (NAISS) at C3SE, partially funded by the Swedish Research
Council through grant agreement no.~2022-06725, and by
the Berzelius resource, provided by the Knut and Alice Wallenberg Foundation at the National Supercomputer Centre.

{\small
    \bibliographystyle{ieeenat_fullname}
    \bibliography{main}}

\end{document}

%% file: preamble.tex
\usepackage{tipa}
\usepackage{todonotes}
\usepackage{multirow}
\usepackage{textcomp}
\usepackage{colortbl}
\usepackage{mathtools}
\usepackage{amssymb}  %
\usepackage{amsfonts}
\usepackage{amsthm}
\usepackage{tabu}
\usepackage{siunitx}
\usepackage{enumitem}

\newtheorem{theorem}{Theorem}[section]

\theoremstyle{definition}
\newtheorem{definition}[theorem]{Definition}

\newcommand{\ours}[0]{SIMFIT++}
\newcommand{\poselib}[0]{\texttt{PoseLib}}
\newcommand{\posebench}[0]{\texttt{PoseBench}}

\newcommand{\magsac}[0]{\texttt{MAGSAC++}}
\newcommand{\homog}[1]{[#1;1]}
\newcommand{\norm}[1]{\left\lVert#1\right\rVert}
\usepackage{todonotes}

\usepackage{algorithm}
\usepackage{algpseudocode}
\usepackage{amsmath}

%% file: sec/abstract.tex
\begin{abstract}
The gold-standard for robustly estimating relative pose through image matching is RANSAC.
While RANSAC is powerful, it requires setting the inlier threshold that determines whether the error of a correspondence under an estimated model is sufficiently small to be included in its consensus set.
Setting this threshold is typically done by hand, and is difficult to tune without a access to ground truth data.
Thus, a method capable of automatically determining the optimal threshold would be desirable.
In this paper we revisit inlier noise scale estimation, which is an attractive approach as the inlier noise scale is linear to the optimal threshold.
We revisit the noise scale estimation method SIMFIT and find bias in the estimate of the noise scale.
In particular, we fix underestimates from using the same data for fitting the model as estimating the inlier noise, and from not taking the threshold itself into account.
Secondly, since the optimal threshold within a scene is approximately constant we propose a multi-pair extension of \ours, by filtering of estimates, which improves results. 
Our approach yields robust performance across a range of thresholds, shown in \Cref{fig:main-results}. Code is available at \url{https://github.com/Parskatt/simfitpp}.
\vfill
\end{abstract}

%% file: sec/intro.tex
\section{Introduction}
\label{sec:intro}
Robust estimation using RANSAC~\cite{fischler1981} requires the user to set a threshold $\tau$, which is used to determine whether a data point should be considered as an inlier to an estimated model or not.
This is in general a difficult problem, partially due to the fact that the optimal threshold can vary significantly between different datasets and models, and is difficult to know without access ground-truth data as is the case in the image matching challenge~\citep{Jin2020} (and also in the real world).

One could therefore wonder whether estimation of the optimal threshold $\tau^*$ could be baked into RANSAC, removing the inconvenience of having to \emph{a priori} know it. 
This is not an entirely novel thought, in fact, there are several approaches that are entirely threshold-free. 
StarSac~\citep{choi2009starsac} performs a grid search over thresholds to find the threshold with lowest model variance, from the insight that a correct threshold produces stable solutions.
However, this is extremely expensive in practice, and the optimal threshold is not guaranteed to align with the lowest model variance. 
In RECON~\citep{raguram2011recon} the problem is reformulated as finding consistency in residual rankings between models, which is independent of the noise level.
ORSA~\citep{moisan2004probabilistic,moisan2012automatic}, finds \emph{meaningful events}~\citep{desolneux2000meaningful} by finding sets of correspondences such that the expectation of their occurrences are as small as possible, which is similarly done in MINPRAN~\citep{stewart1995minpran}, although for probability rather than expectation.
Another of work was developed in ASSC~\citep{wang2004robust} whereby the RANSAC objective is divided by a robust estimate of the scale. 
This objective is however not well-motivated statistically.
ASKC~\citep{wang2009generalized} uses a more well-motivated objective, but still faces challenges with estimating scale from the full contaminated set of correspondences.
In general, these approaches aim to replace RANSAC.
While interesting, in this work we do not wish to get rid of RANSAC, instead we aim to augment it with inlier threshold robustness.
Besides this reason, these approaches have already been shown to under-perform compared to threshold-robust RANSAC approaches on relative pose estimation~\citep{heinrich2013efficient,torr1997development,raguram2011recon}.

Making RANSAC robust to the inlier threshold has also been explored.
MAGSAC and MAGSAC++~\citep{barath2019magsac, barath2020magsac++} marginalize over noise scales, in order to avoid setting a specific threshold. 
Unfortunately, they still need to set an upper threshold, and the choice of this threshold affects results, as we will later show in~\Cref{fig:F-magsac} (our method will be shown to help theirs). 
\citet{hedborg2013fast} estimate the noise level using the forward-backward tracking error, although the algorithm used is not specified.

Closest to our work is SIMFIT~\citep{heinrich2013efficient} which alternates between estimating the inlier noise scale $\sigma$ and running MSAC~\citep{torr2000mlesac} using a threshold computed using $\sigma$.
This approach forms the basis of our work.
Note that, while we compare ours mainly to SIMFIT, SIMFIT has already been shown to yield SotA results~\citep{heinrich2013efficient}.
Although SIMFIT produces good results, we find that their estimate is biased, due to estimating the variance on the fitting set, and also due to not taking the inlier threshold into account.
We propose two fixes to this problem.
First, we perform a train/validation split for the estimation of the geometric model and the inlier noise scale.
Secondly, we propose an adjustment to the noise scale estimator which works also for thresholded distributions.
We call this algorithm \ours.
Finally, we propose a multi-pair extension of \ours~by filtering estimates over larger image-sets.
This multi-pair extension further improves results, particularly when the initial threshold provided is far from optimal.
\paragraph{Our main contributions are:}
\begin{enumerate}[label=\textbf{\alph*)}]
    \item We propose an empirically well-motivated estimator of the noise scale $\sigma^*$, by identifying two sources of bias in previous approaches, and integrate it into a RANSAC framework. This is described in~\Cref{sec:less-biased}.
    \item We extend the estimator to a multi-pair setting in~\Cref{sec:multi-pair}.
    \item We conduct large-scale evaluations and ablations, which show that our approach works well in practice in~\Cref{sec:results}. 
\end{enumerate}

%% file: sec/method.tex
\section{Method}
\label{sec:method}
We present \ours~in \Cref{alg:ours}.
In this paper we focus only on relative pose estimation, but the algorithm can be easily generalized, as done in SIMFIT, to cover other types of geometric estimation problems.
Our algorithm is similar to SIMFIT, but differs in some key areas.
In the next sections we go into more details of our proposed method.
\subsection{Modeling the empirical distribution of inliers}
A classic assumption in multi-view computer vision is that the detection error is Normally distributed, and by propagation find that the squared residuals (when estimating relative pose) ought to be $\chi_k^2$ distributed, where $k$ depends on the dimensionality of the model/residuals. 
Typically the assumption is that $k=1$.
Next, we briefly recap when this is true.
\paragraph{When are the squared residuals theoretically $\chi_1^2$?}
Assuming that our detections $\mathbf{x}_A, \mathbf{x}_B\in\mathbb{R}^2$ are affected by independent Normally distributed noise, 
\begin{equation}
\label{eq:det-noise}
\mathbf{x}_A = \mathbf{x}_A^* + \mathbf{\varepsilon}_A, \mathbf{x}_B = \mathbf{x}_B^* + \mathbf{\varepsilon}_B, \varepsilon \sim \mathcal{N}(0,\sigma^2\mathbf{I})
\end{equation}
where $\mathbf{x}_A^* = \pi(\mathbf{x}^*), \mathbf{x}_B^* = \pi(\mathbf{R} \mathbf{x}^* +\mathbf{t}), \pi: \mathbb{R}^3\to \mathbb{R}^2$. Of course, if $\mathbf{x}^*$ was known, the sum of squared residuals would be distributed as
\begin{equation}
\norm{\mathbf{x}_A-\mathbf{x}_A^*}^2 + \norm{\mathbf{x}_B-\mathbf{x}_B^*}^2 \sim \chi^2_4.
\end{equation}
However, since $\mathbf{x}^*$ is unknown, we try to find the most likely $\mathbf{x}$, through
\begin{align}
    \min_{\hat{\mathbf{x}}_A, \hat{\mathbf{x}}_B} \norm{\mathbf{x}_A-\hat{\mathbf{x}}_A}^2 + \norm{\mathbf{x}_B-\hat{\mathbf{x}}_B}^2 \\
    \text{s.t.}\quad {\hat{\mathbf{x}}_B}^{\top} \mathbf{E} \hat{\mathbf{x}}_A = 0
\end{align}
The Sampson~\citep{sampson1982fitting} error for the Fundamental matrix~\citep{luong1996fundamental} can be derived by linearizing the constraint and solving a Lagrangian, and is often very close to the optimal solution~\citep{rydell2024revisiting}. It is given by
\begin{equation}
    \varepsilon^2_S = \frac{([\mathbf{x}_B;1]^{\top} \mathbf{E} [\mathbf{x}_A;1])^2}{\norm{\mathbf{E}_{12}[\mathbf{x}_A;1]}^2 + \norm{(\mathbf{E}^{\top})_{12}[\mathbf{x}_B;1]}^2}.
\end{equation}
We will consider the signed Sampson error as
\begin{equation}
    \varepsilon_S = \text{sign}([\mathbf{x}_B;1]^{\top} \mathbf{E} [\mathbf{x}_A;1])\sqrt{\varepsilon^2_S}.
\end{equation}
Now, we investigate $\varepsilon_S$.
Inserting \Cref{eq:det-noise} into the numerator and expanding we get
\begin{equation}
    \homog{\mathbf{x}_B}^{\top} \mathbf{E} \homog{\mathbf{x}_A} = \homog{\mathbf{x}^*_B+\varepsilon_B}^{\top}\mathbf{E} \homog{\mathbf{x}^*_A+\varepsilon_A},
\end{equation}
further expanding the right hand side
\begin{equation}
\homog{\mathbf{x}^*_B}^{\top}\mathbf{E} [\varepsilon_A;0] + [\varepsilon_B;0]^{\top}\mathbf{E} [\varepsilon_A;0] + \homog{\mathbf{x}^*_A}^{\top}\mathbf{E}^{\top}[\varepsilon_B;0]
\end{equation}
and ignoring the quadratic terms we have
\begin{equation}
\homog{\mathbf{x}^*_B}^{\top}\mathbf{E} [\varepsilon_A;0] + \homog{\mathbf{x}^*_A}^{\top}\mathbf{E}^{\top} [\varepsilon_B;0] \sim \mathcal{N}(0, s^2)
\end{equation}
where we used the fact that sums of Normals are Normal.
The variance $s^2$ is
\begin{equation}
    s^2 = \sigma^2(\norm{\mathbf{E}_{12}[\mathbf{x}^*_A;1]}^2 + \norm{(\mathbf{E}^{\top})_{12}[\mathbf{x}^*_B;1]}^2).
\end{equation}
For sufficiently small $\varepsilon$ we can consider the denominator as constant
\begin{align}
    \sqrt{\norm{\mathbf{E}_{12}[\mathbf{x}_A;1]}^2 + \norm{(\mathbf{E}^{\top})_{12}[\mathbf{x}_B;1]}^2} \approx \\
    \sqrt{\norm{\mathbf{E}_{12}[\mathbf{x}^*_A;1]}^2 + \norm{(\mathbf{E}^{\top})_{12}[\mathbf{x}^*_B;1]}^2}
\end{align}
Hence,
\begin{equation}
    \varepsilon_S \sim \mathcal{N}(0,\sigma^2), \varepsilon^2_S/\sigma^2\sim \chi_1^2.
\end{equation}

It is worth considering the main assumptions for this analysis. First, we assume that the detection errors are Normal and independent. For detector based methods this seems reasonable, but for others, like dense matchers, they seem less likely.
Secondly, we assume that the inlier noise is small compared to the observations. This is likely reasonable in practice.

\paragraph{Are the squared residuals $\chi_1^2$ in practice?}
While the above analysis shows that the Sampson error is Normal if we assume that the detection noise is Normal, it does not answer whether the detection noise is Normal.

\begin{figure}
    \centering
    \includegraphics[width=\linewidth]{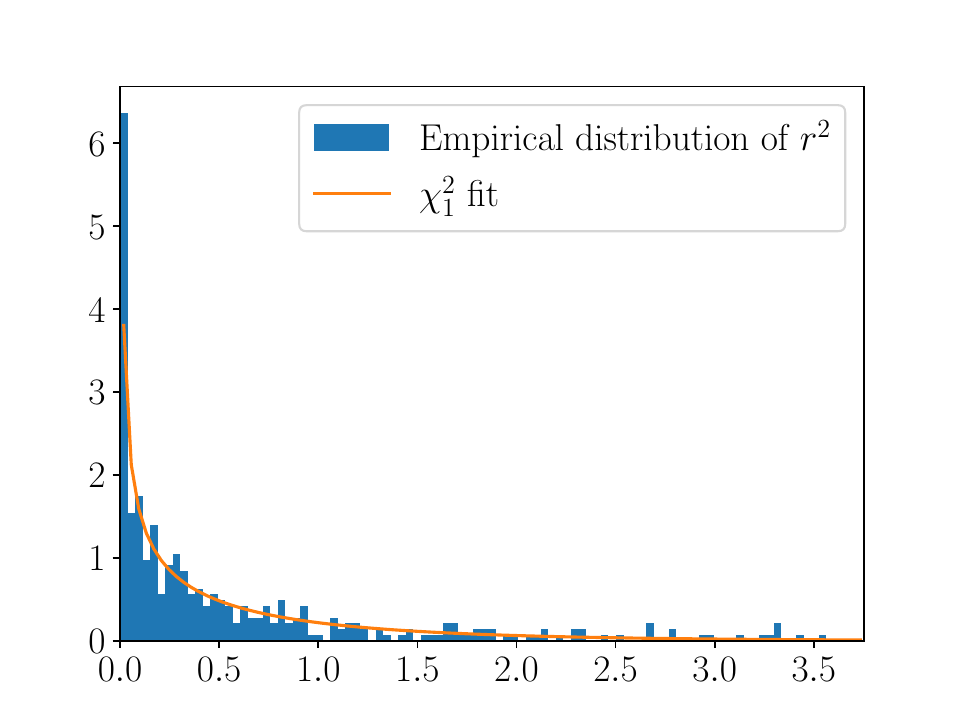}
    \caption{\textbf{Qualitative comparison of fit to empirical residual distribution for SuperGlue.} We show the distribution of squared residual errors for SuperPoint~\citep{detone2018superpoint} with SuperGlue~\citep{sarlin2020superglue} on a randomly selected pair in the MegaDepth1500~\citep{sun2021loftr} test set.}
    \label{fig:hist-fit-superpoint}
\end{figure}

\begin{figure}
    \centering
    \includegraphics[width=\linewidth]{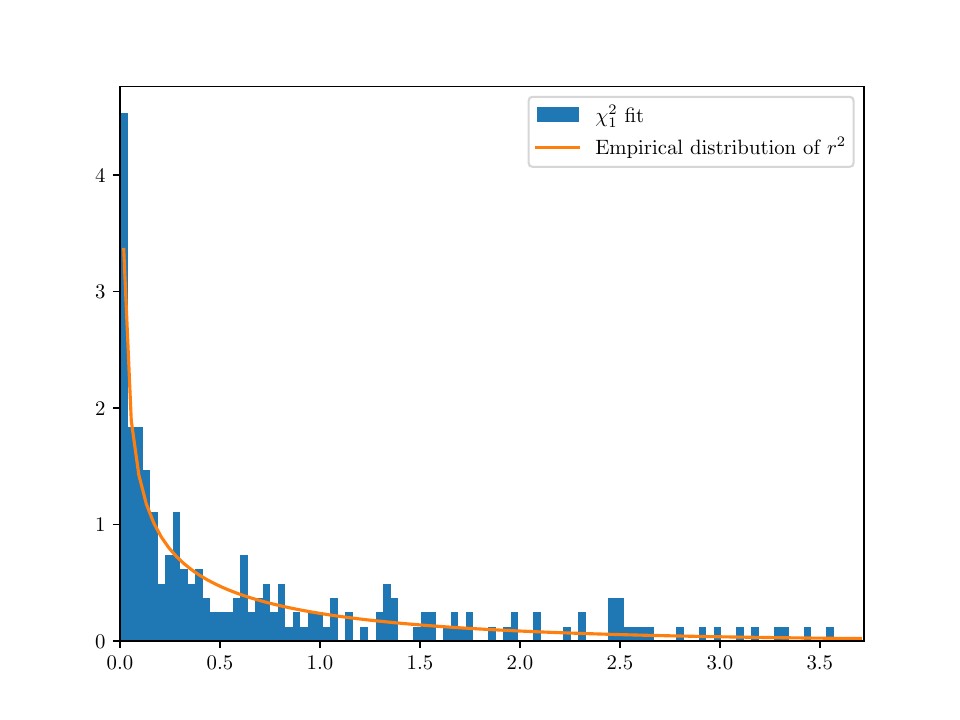}
    \caption{\textbf{Qualitative comparison of fit to empirical residual distribution for SuperGlue.} We show the distribution of squared residual errors for SuperPoint~\citep{detone2018superpoint} with SuperGlue~\citep{sarlin2020superglue} on a randomly selected pair in the ScanNet~\citep{dai2017scannet} test set.}
    \label{fig:hist-fit-superpoint-scannet}
\end{figure}

\begin{figure}
    \centering
    \includegraphics[width=\linewidth]{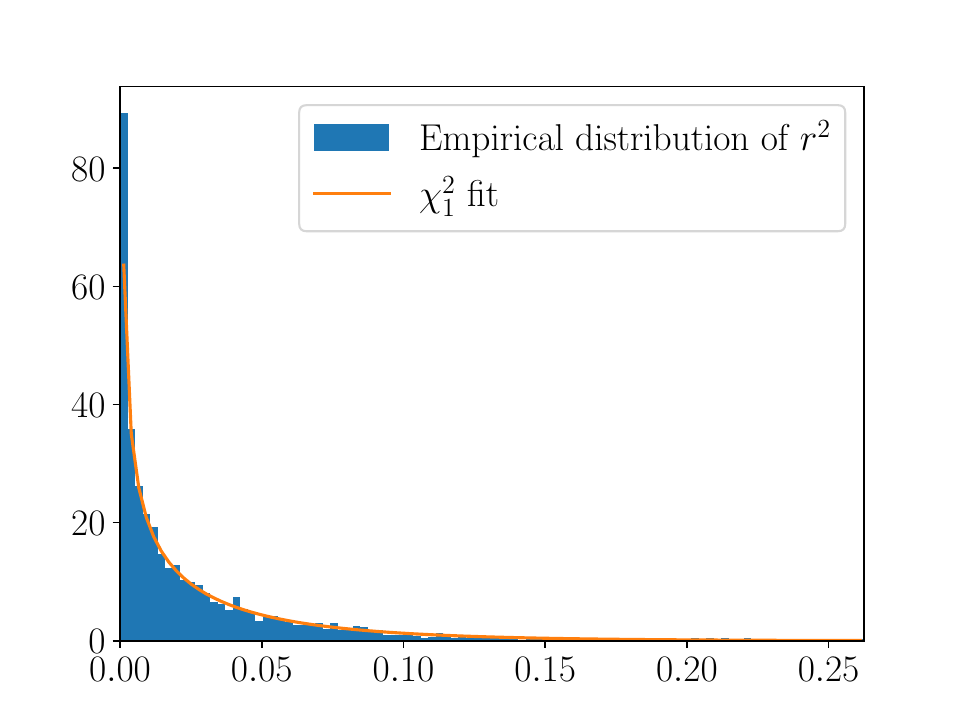}
    \caption{\textbf{Qualitative comparison of fit to empirical residual distribution for RoMa.} We show the distribution of squared residual errors for RoMa~\citep{edstedt2024roma} on a randomly selected pair in the MegaDepth1500~\citep{sun2021loftr} test set.}
    \label{fig:hist-fit-roma}
\end{figure}

\begin{figure}
    \centering
    \includegraphics[width=\linewidth]{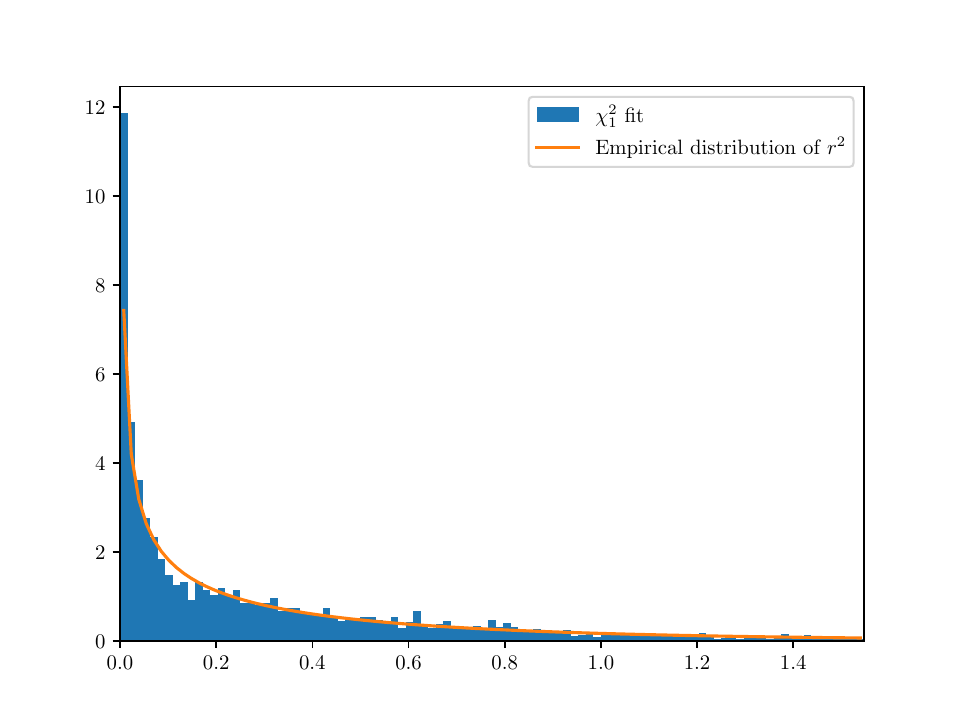}
    \caption{\textbf{Qualitative comparison of fit to empirical residual distribution for RoMa.} We show the distribution of squared residual errors for RoMa~\citep{edstedt2024roma} on a randomly selected pair in the ScanNet~\citep{dai2017scannet} test set.}
    \label{fig:hist-fit-roma-scannet}
\end{figure}

To this end there is not much we can do except some qualitative analysis of different detectors and matchers.
We start by looking at a qualitative example of SuperGlue correspondences from SuperPoint detections in~\Cref{fig:hist-fit-superpoint,fig:hist-fit-superpoint-scannet}. 
As can be seen in the figure, the $\chi_1^2$ assumption seems to hold quite well in this case.
We next have a look at matches from the dense image matcher RoMa in~\Cref{fig:hist-fit-roma,fig:hist-fit-roma-scannet}.
Somewhat surpisingly, the assumption still seem to hold.
Note however, that we do not prove that detections are i.i.d. Normal, but only that the resulting squared residuals are $\chi_1^2$.
In conclusion, the $\chi_1^2$ model holds pretty well for correspondences, which is lucky, as our life would otherwise be significantly more difficult\footnote{We apologize for the informal phrasing.}.

\subsection{Threshold $\tau$ from $\sigma$.}
When $\sigma$ is known, the threshold $\tau$ is typically assumed to only depend on the confidence level $\alpha$ through
\begin{equation}
    \tau^* = F^{-1}(\alpha)\sigma^*,
\end{equation}
where $F^{-1}$ is the inverse CDF (the quantile function) of the inlier noise distribution. 
If $\alpha$ is set to, \eg $0.95$, it can be interpreted~\citep[p. 119]{hartley2003multiple} as that the probability that an inlier will be incorrectly rejected is $5\%$, given that the model hypothesis (including both the estimated geometric model and the model of the inlier distribution) is true.
An obvious fact which nevertheless should be noted is that under different assumed inlier distributions the scaling between $\tau$ and $\sigma$ can differ significantly for the same $\alpha$.

\subsection{A Less Biased Estimator of $\sigma$}
\label{sec:less-biased}
Let's first recall the definition of an unbiased estimator
\begin{definition}
    Let $\hat{\theta} = f(x)$ be an estimator of a parameter $\theta$, and $p(x|\theta)$ be the data distribution conditioned on $\theta$.
    Then $f$ is said to be unbiased if and only if 
    \begin{equation}
        \mathbb{E}_{x\sim p(x|\theta)}[f(x)-\theta] = 0, \forall\theta
    \end{equation}
\end{definition}
It might seem quite easy to make an unbiased estimator, but even for simple models many seemingingly reasonable estimators are biased.
Famously the estimator 
\begin{equation}
    \hat{\sigma}^2(x) = \frac{1}{N}\sum_{i=1}^N(x_i-\frac{1}{N}\sum_{j=1}^Nx_j)^2
\end{equation}
 is an underestimate (with the Bessel correction $\frac{N}{N-1}$ for $N>1$).
 The bias typically comes from the fact that intermediate results in the estimator are random variables. 
 While this bias can be expressed analytically as for the case above, for more complex models this is difficult or impossible.
 In our case the intermediate result is the estimate of the geometric model, which is a random variable of the full RANSAC pipeline.
 While it is in principle possible to express this uncertainty~\citep{kanatani2012renormalization}, it would make our method significantly more complex.

 Opting away from this complexity, we will instead turn our attention to low-hanging fruit. 
 We begin by introducing the previous noise scale estimator SIMFIT~\citep{heinrich2013efficient} in~\Cref{alg:simfit}. 
 SIMFIT repeatedly runs RANSAC (specifically MSAC) on a shrinking inlier set, and estimates the inlier noise scale $\sigma$ on the inlier set with the median squared residual estimator~\citep{rousseeuw1987robust,torr1997development,heinrich2013efficient}
\begin{equation}
    \hat{\sigma}^2 = \frac{\text{median}(r^2)}{F^{-1}(0.5)}.
\end{equation}
The denominator tells us where we expect the value of the $50^\text{th}$ percentile residual to be, while the nominator is our empirical estimate. The fraction is then our scale. This estimator has several nice properties. It is asymptotically consistent~\citep{heinrich2013efficient}, and is quite robust with a $50\%$ breakdown rate~\citep{rousseeuw1987robust}. However, it does not account for the dependencies between $r^2$ and the estimated model $\mathcal{H}$ nor the fact that we only include residuals smaller than our threshold $\tau^2$.
\begin{algorithm}
\caption{SIMFIT~\citep{heinrich2013efficient}.}
\label{alg:simfit}
\begin{algorithmic}[1]
\Require{$\mathbf{m}\in \mathbb{R}^{N\times 2 \times 2}$: Matches. $\tau_0$: Threshold guess. $\alpha$: confidence level.}
\State $\tau^* \gets \tau_0$ \Comment{Init threshold with best guess.}
\State $\text{inliers} \gets \{1,\dots,|\mathbf{m}|\}$
\Repeat
    \State $\mathcal{H}, \mathbf{r}^2$ $\gets$ RANSAC$(\mathbf{m}_{\text{inliers}}, \tau^*)$
    \State $\text{inliers} \gets \{\text{inliers}_j | \mathbf{r}_j^2 \le {\tau^*}^2, 1\le j\le |\text{inliers}|\}$
    
    \State $\hat{\sigma} \gets$ \Call{MedianEstimator}{$\mathbf{r}_{\text{inliers}}^2$}
    \State $\tau^* \gets$ $F^{-1}(\alpha)\hat{\sigma}$
\Until{$\tau^*$ converged}
\State $\text{inliers} \gets$ \Call{ModelShift}{$\mathcal{H}, \mathbf{m}, \text{inliers}$} \Comment{Optional.}
\State $\mathbf{r}_{\text{inliers}}^2\gets$ \Call{ComputeResiduals}{$\mathcal{H}, \mathbf{m}_{\text{inliers}}$}
\State $\hat{\sigma} \gets$ \Call{MedianEstimator}{$\mathbf{r}_{\text{inliers}}^2$}
\State \Return $\tau^*$
\end{algorithmic}
\end{algorithm}

We present our proposed estimator, \ours, which reduces these biases, in~\Cref{alg:ours}.
\ours~iteratively improves the estimate of $\tau$ by drawing subsets of the matches $\mathbf{m}$, with proportions $p$ for the estimation of the geometric model, and $1-p$ for the estimation of the inlier noise scale.
At each iteration $\sigma$ is estimated with a debiased median estimator, and subsequently filtered using the previous estimate.
This process is repeated, either until a tolerance criterion is reached (ftol $=0.01$), or for a maximum of $4$ iterations.
We next explain the motivation for our method in more detail.
\begin{algorithm}
\caption{\ours~(Ours).}
\label{alg:ours}
\begin{algorithmic}[1]
\Require{$\mathbf{m}\in \mathbb{R}^{N\times 2 \times 2}$: Matches. $\tau_0$: Threshold guess. $\tau_{\min}$: Lowest allowed $\tau$. Default $1/4$ pixels. $\tau_{\max}$: Highest allowed $\tau$. Default $8$ pixels. $\alpha$: confidence level. $p$: Train split proportion.}
\State $\tau^* \gets \tau_0$ \Comment{Init threshold with best guess.}
\Repeat
    \State Draw train, val set with proportions $p, 1-p$. 
    \State $\mathcal{H}, \mathbf{r}_{\text{train}}^2$ $\gets$ RANSAC$(\mathbf{m}_{\text{train}}, \tau^*)$
    \State $\hat{\sigma}\gets$ \Call{$\tau$-MedianEstimator}{$\mathbf{r}_{\text{val}}^2,\tau^*$}
    \State $\tau^* \gets$ \Call{filter}{$\tau^*, F^{-1}(\alpha)\hat{\sigma}$}
\Until{$\tau^*$ converged}
\State \Return $\tau^*$
\end{algorithmic}
\end{algorithm}

\paragraph{Reducing bias due to overfitting:}
Our first contribution is to use separate training and validation sets for the estimation of the geometric model and the inlier noise scale.
This is closely related to cross-validation and ensembling, which are well-known statistical techniques to reduce bias and variance.
Splitting into training and validation reduces much of the issues with RANSAC overfitting to the training set~\citep{brissman2023camera}, which, if not adjusted for, produces an underestimate of the inlier noise scale.
The choice of $p$ in~\Cref{alg:ours} we fix as $0.5$.
If the number of matches is very low it would likely be better to use a larger $p$, and for extreme cases leave-one-out cross-valation.
We leave such investigations as future work.

\begin{figure}
    \centering
    \includegraphics[width=\linewidth]{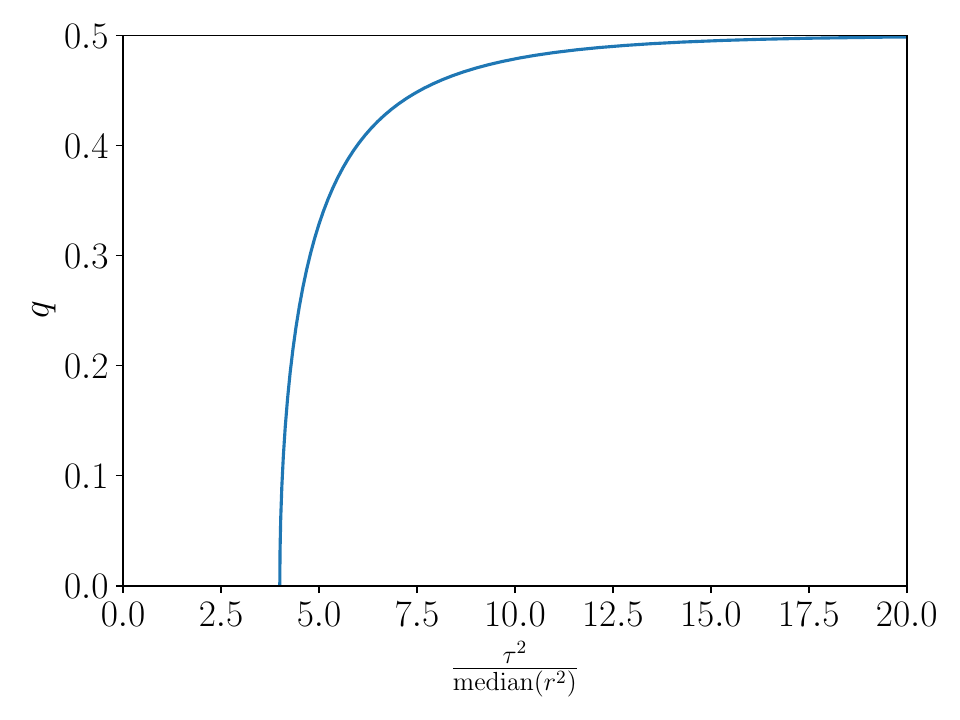}
    \caption{\textbf{Corrected percentile depending on threshold and median squared residual.} Not taking the threshold into account is equivalent to setting $q=0.5$.}
    \label{fig:corrected}
\end{figure}

\paragraph{Reducing bias due to $\tau$:}
The median estimator, as used in SIMFIT, will be biased when the residuals come from a truncated $\chi_1^2$, and not from an untruncated $\chi_1^2$, as assumed.
To create an unbiased estimate, we need to adjust the percentile in the denominator accordingly.
A simple way is to introduce a variable $q\in[0,0.5]$ which compensates for this bias, i.e.,
\begin{equation}
    \hat{\sigma}^2 = \frac{\text{median}(r^2)}{F^{-1}(q)}.
\end{equation}
Which leaves us with the task of estimating $q$.
We approach this by setting $q_0=0.5$ and 
\begin{align}
q_k &= F(\frac{\tau^2}{\hat{\sigma}_{k-1}^2})/2, \\
\sigma_k^2 &= \frac{\text{median}(r^2)}{F^{-1}(q_k)}.
\end{align}
This gives us a sequence of $\sigma$ that quickly converges, in about 5 iterations, in practice. 
It should be noted that this sequence converges for $\tau^2\text{median}(r^2)^{-1} > \approx 4$. 
This can be shown by finding the fix-point
\begin{align}
    q &= F\left(\frac{\tau^2}{\sigma^2}\right)/2 = F\left(\frac{\tau^2 F^{-1}(q)}{\text{median}(r^2)}\right)/2 \\
    &\Leftrightarrow F^{-1}(2q) = \frac{\tau^2}{\text{median}(r^2)}F^{-1}(q) \\
    &\Leftrightarrow \frac{F^{-1}(2q)}{F^{-1}(q)} = \frac{\tau^2}{\text{median}(r^2)}
\end{align}
We plot this relation in~\Cref{fig:corrected}.

\paragraph{Reducing variance by filtering:}
Finally, in contrast to SIMFIT, which uses the final $\tau^*$ as estimate, we average our estimate across the runs, and include an upper and lower threshold, which can be seen as a uniform prior belief on the minimum and maximum possible thresholds. 
In practice this filter is an online mean filter, which updates only when the estimates threshold is within the prior belief bounds.

\subsection{Extension to the Multi-Pair Setting.}
\label{sec:multi-pair}
\begin{figure*}
    \includegraphics[width=.32\linewidth]{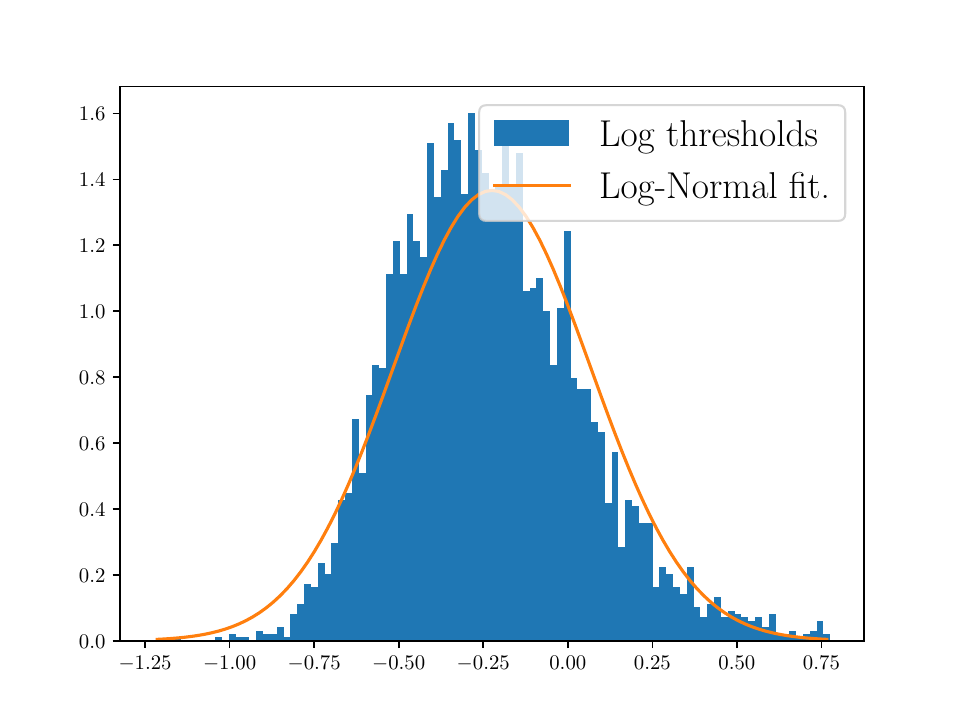}
    \includegraphics[width=.32\linewidth]{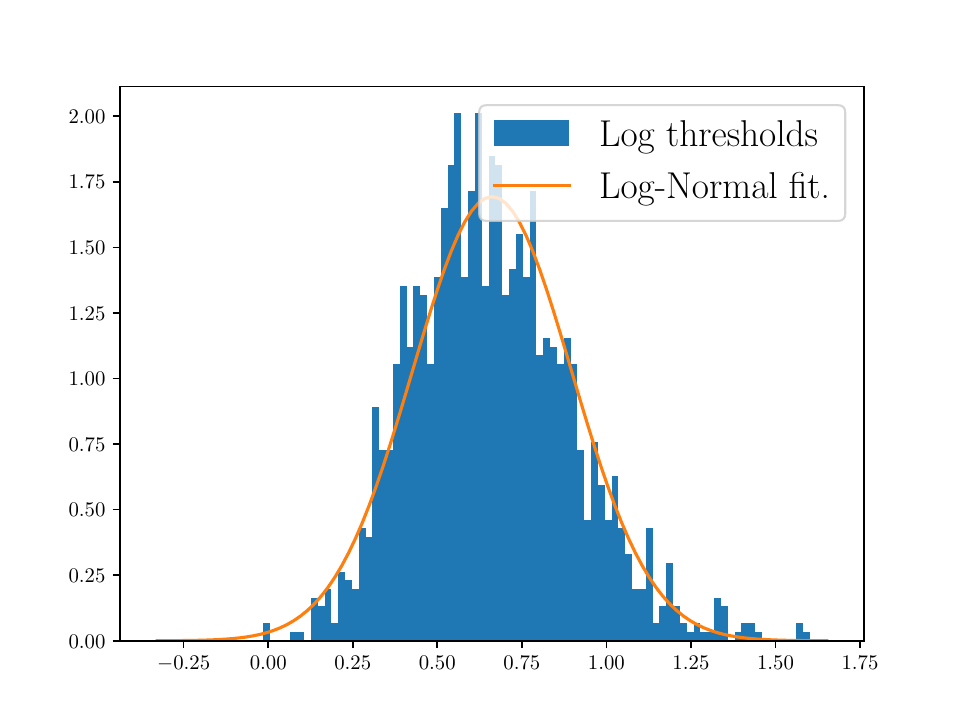}
    \includegraphics[width=.32\linewidth]{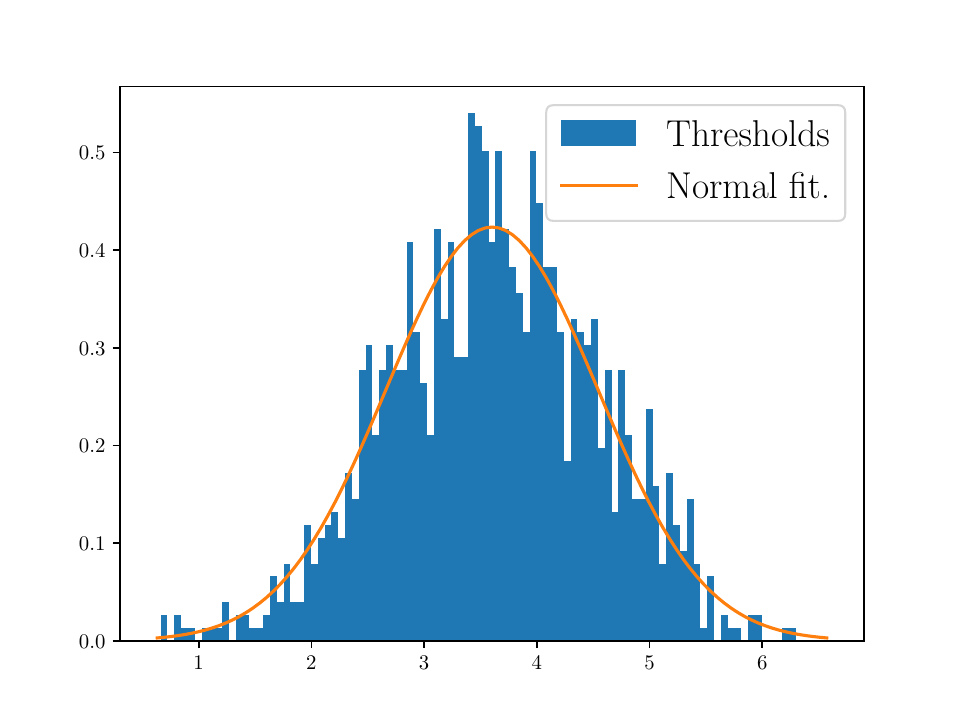}
    \caption{\textbf{Estimated threshold distributions.} We run SuperPoint+SuperGlue over three different dataset with \ours~and investigate the resulting distribution of estimated thresholds. \textbf{Left/Middle:} Distribution for the ``British Museum'' and the MegaDepth1500 scenes. Here a Log-Normal assumption seems warranted. \textbf{Right:} Distribution for ScanNet1500. Here it instead seems as if the distribution is Normal. In our multi-pair filtering we use an online geoemetric mean estimator, which is well-motivated if the distribution is unimodal symmetric in log-space.}
    \label{fig:thr-dist}
\end{figure*}

Extending the single-pair algorithm to multiple pairs was first suggested, to the best of our knowledge, by Torr: \emph{``The standard deviation can be estimated between each pair of images and the results filtered over time.''}~\citep{torr1997development}.
In~\Cref{alg:multi} we provide a practical implementation of this idea.
Our algorithm iteratively updates the threshold $\tau$ through filtering estimates from single pairs.
The filtering operation is similar to the one in the single pair algorithm, with the difference that we use the geometric mean instead of the arithmetic mean, as we empirically observed that the distribution seems to be log Normal, rather than Normal.
To validate this assumption we show some qualitative results from running SuperGlue on a few different scenes~\Cref{fig:thr-dist}.
A possible reason for why ScanNet1500 shows Normal behaviour instead of log-Normal is that the images all have a similar zoom, while Phototourism-type images typically contains different zoom-levels.

\begin{algorithm}
\caption{\ours~multi.}
\label{alg:multi}
\begin{algorithmic}[1]
\Require{$\mathcal{D}$: Dataset of matches. $\tau_0$: Threshold guess.  $\tau_{\min}$: Lowest allowed $\tau$. Default $1/4$ pixels. $\tau_{\max}$: Highest allowed $\tau$. Default $8$ pixels. $\alpha$: confidence level. $p$: Train split proportion.}
\State $\tau^*_{\text{multi}} \gets \tau_0$ \Comment{Init threshold with best guess.}
\Repeat
    \State $\mathbf{m} \sim \mathcal{D}$ \Comment{Draw matches from dataset.}
    \State $\tau^* \gets$ \Call{\ours}{$\mathbf{m}, \alpha, \tau^*, \tau_{\min}, \tau_{\max}, p$}
    \State $\tau^*_{\text{multi}} \gets$ \Call{filter}{$\tau^*_{\text{multi}}, \tau^*$}
\Until{$\tau^*_{\text{multi}}$ converged}
\State \Return $\tau^*_{\text{multi}}$
\end{algorithmic}
\end{algorithm}

%% file: sec/results.tex
\section{Results}
\label{sec:results}

\subsection{Experimental details}
We fix the number of RANSAC iterations to $500$ across experiments. While running further iterations gave slight increases in performance, the optimal threshold was the same, and we thus use a more reasonable number of iterations for computational reasons. 
We set the number of \ours~iterations as $N=4$. 
We set $\alpha=0.99$ for all experiments, consider $\tau_0 \in [0.5, 4]$, and set $\tau_{\min} = 0.25, \tau_{\max} = 8$, and $p=0.5$.

We compare \ours~with fixed threshold RANSAC, as well as SIMFIT. 
For fair comparison, after estimating $\sigma$ we run a final RANSAC from scratch.
We mainly consider the robust estimators in \poselib, but also the OpenCV implementation of \magsac\footnote{\url{github.com/opencv/opencv}}. For \magsac~ we additionally refine the RANSAC model with \poselib, as we found it to yield more better results. 
As a side-note, we empirically find that using post RANSAC refinement causes the optimal threshold for the OpenCV \magsac~to shift significantly upwards, which might explain the common belief that this specific RANSAC requires an extremely low threshold\footnote{See for example \url{https://ducha-aiki.github.io/wide-baseline-stereo-blog/2021/05/17/OpenCV-New-RANSACs.html}.} (however, we do find that the optimal threshold is somewhat lower than for \poselib).

\subsection{Datasets}
We use \texttt{posebench}\footnote{\url{github.com/vlarsson/posebench}
} for all matching data. For relative pose this includes the IMC-PT~\citep{jin2021image}, MegaDepth1500~\citep{sun2021loftr}, and ScanNet1500~\citep{sarlin20superglue}  datasets. The feature matchers SIFT~\citep{lowe2004distinctive} SuperGlue~\cite{sarlin20superglue}, LightGlue~\citep{lindenberger2023lightglue}, ASpanFormer~\citep{chen2022aspanformer}, DKM~\citep{edstedt2023dkm}, and RoMa~\citep{edstedt2024roma} are represented. This makes it a suitable benchmark for matching, as it covers a large set of problems and modern as well as classical feature matching methods.

\subsection{Essential Matrix Estimation}
\begin{figure}
    \centering
    \includegraphics[width=\linewidth]{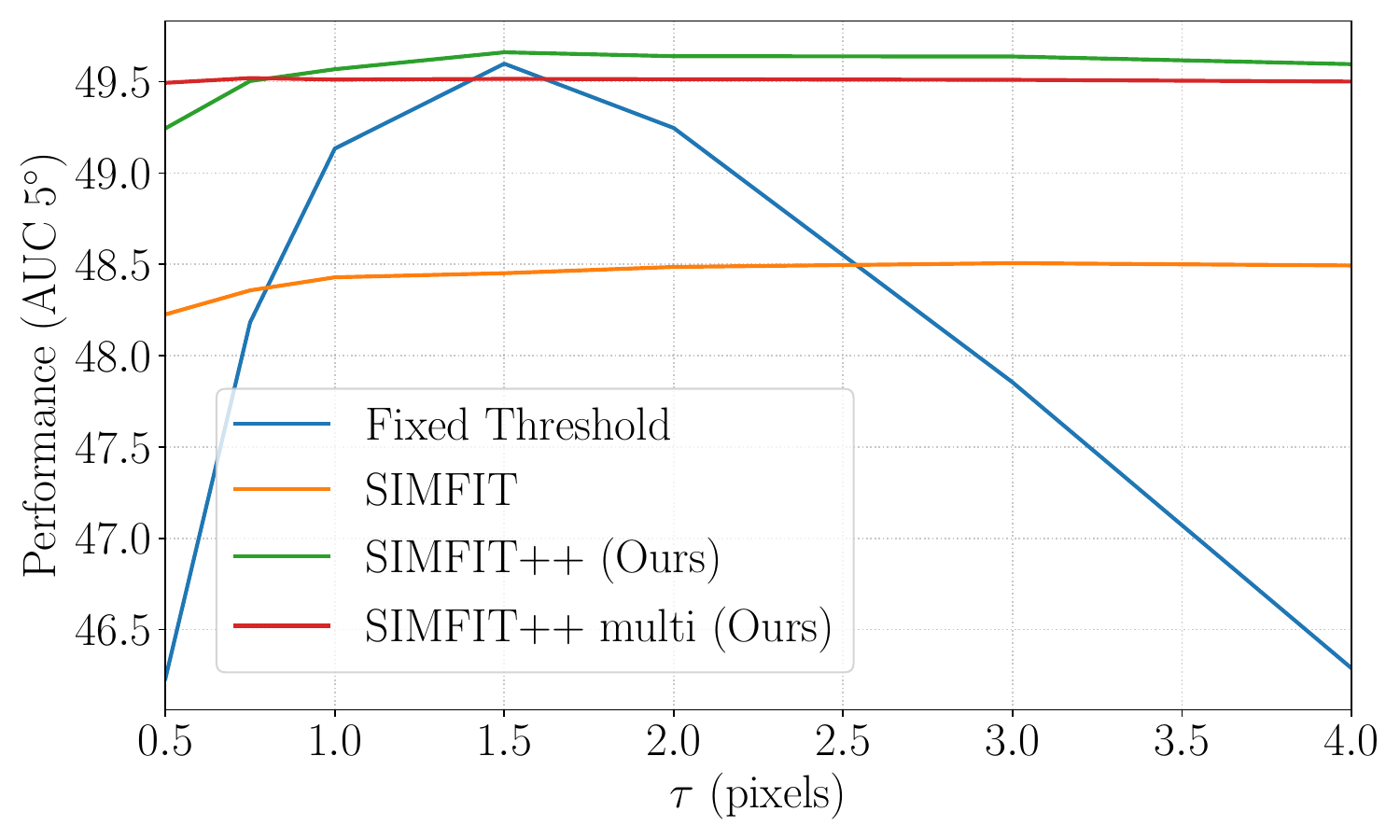}
    \caption{\textbf{Essential matrix estimation performance as a function of the threshold $\tau$ using \poselib.}}
    \label{fig:E-poselib}
\end{figure}

We investigate results for estimation of Essential matrix using \poselib. 
Results are presented in \Cref{fig:E-poselib}, and in \Cref{tab:E-poselib}.
As can be seen, \ours, outperforms both all fixed thresholds, and the previous method SIMFIT.
\input{tables/E-poselib}

\subsection{Fundamental Matrix Estimation}
\begin{figure}
    \centering
    \includegraphics[width=\linewidth]{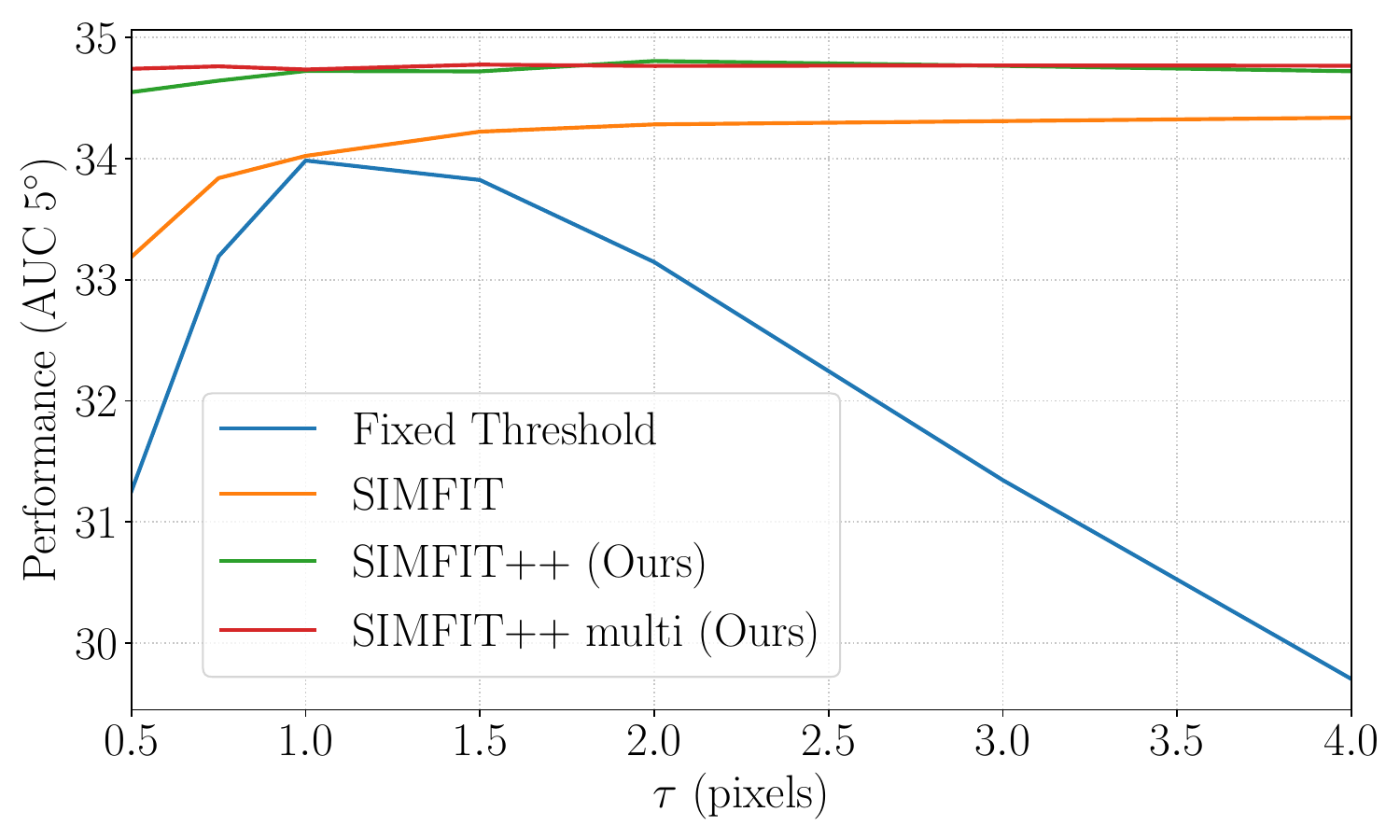}
    \caption{\textbf{Fundamental matrix estimation performance as a function of the threshold $\tau$ using \poselib.}}
    \label{fig:F-poselib}
\end{figure}

The Fundamental matrix captures the epipolar geometry between two \emph{uncalibrated} pinhole camera images.
Compared to the Essential matrix, it has more degenerate configurations, this means that diverse correspondences are more important.
On this task we compare both \poselib~and \magsac.
\paragraph{\poselib:}
We present results for this experiment in~\Cref{fig:F-poselib}, with a detailed table presented in~\Cref{tab:F-poselib}.
\input{tables/F-poselib}
Here our improvements compared to the best fixed threshold is more pronounced than for Essential matrix estimation.
Note additionally that the optimal fixed threshold seems to be significantly lower for Fundamental matrix estimation, possibly indicating that precision is more important than recall in this case, \ie, more important to avoid outliers.

\begin{figure}
    \centering
    \includegraphics[width=\linewidth]{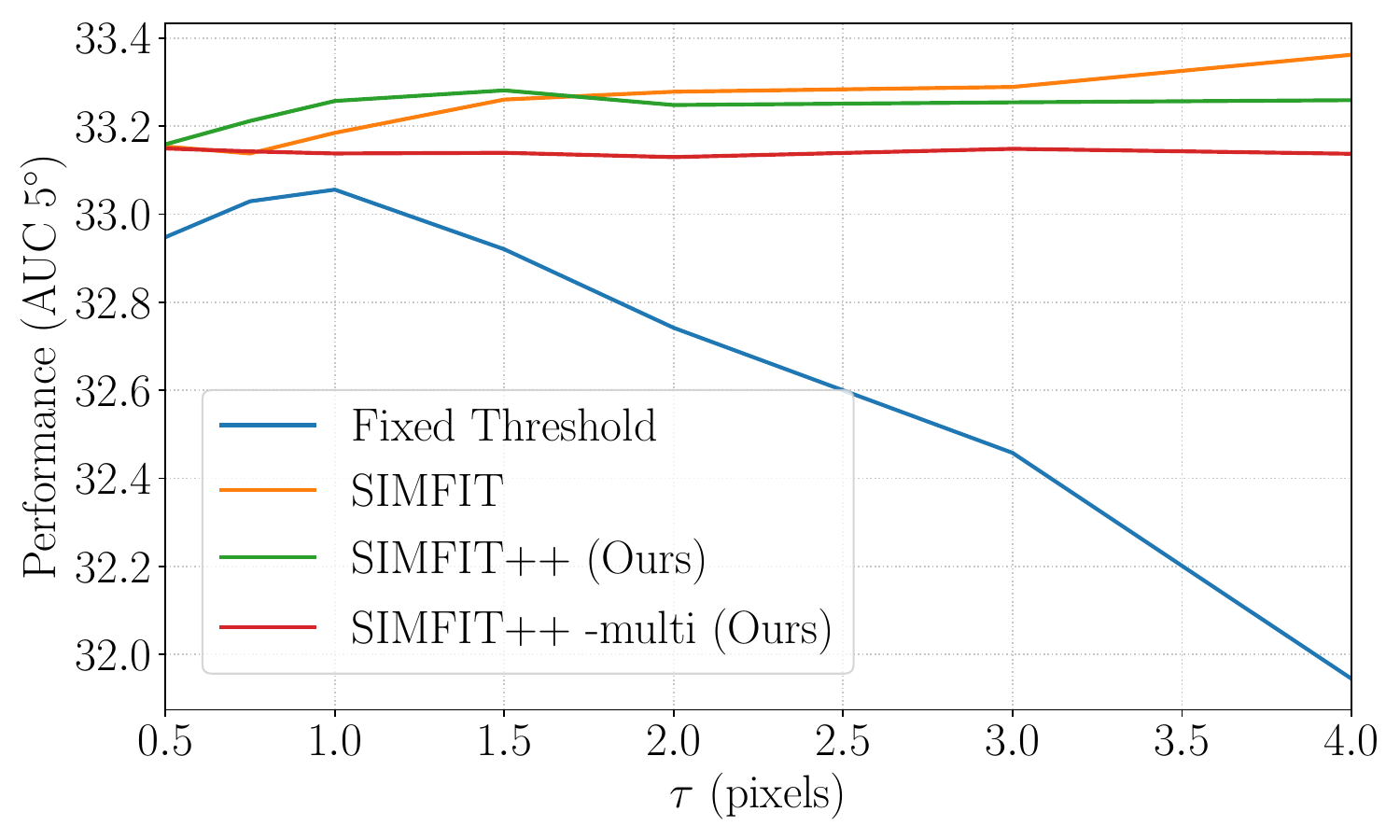}
    \caption{\textbf{Fundamental matrix estimation performance as a function of the threshold $\tau$ using \magsac.} Note that the range of the performance is tighter, as \magsac~is significantly more robust to the threshold than~\poselib.}
    \label{fig:F-magsac}
\end{figure}

\paragraph{\magsac:}
We present results for this experiment in~\Cref{fig:F-magsac}.
Perhaps as expected, the performance is significantly more stable across thresholds for \magsac ~compared to \poselib, however the performance for the optimal threshold is in general lower than for \poselib.
This is expected, as \magsac~marginalizes over noise levels within its threshold range.

\subsection{Per-Dataset Threshold Oracle Comparison}
\begin{table}
  \centering
  \caption{\textbf{Relative decrease in Essential matrix estimation performance compared to oracle thresholds.} We investigate how our estimator compares to using an oracle, selecting the best performing threshold based on ground truth data per dataset.}
  \begin{tabular}{lccc}
    \toprule
      & Mega1500 & ScanNet & IMC-PT \\
    \midrule
    SIMFIT & -2.7 & -0.6 & -2.0 \\
    \ours & -0.5 & -0.3 & -1.2 \\
    \ours~multi & -0.6 & -0.2 & -0.9 \\
    \bottomrule
  \end{tabular}
  \label{tab:oracle-E}
\end{table}
\begin{table}
  \centering
  \caption{\textbf{Relative decrease in Fundamental matrix estimation performance compared to oracle thresholds.} We investigate how our estimator compares to using an oracle, selecting the best performing threshold based on ground truth data per dataset.}
  \begin{tabular}{lccc}
    \toprule
      & Mega1500 & ScanNet & IMC-PT \\
    \midrule
    SIMFIT & -0.8 & -0.8 & -1.4 \\
    \ours& -0.4 & -0.2 & -0.5 \\
    \ours~multi & -0.6 & -0.1 & -0.3 \\
    \bottomrule
  \end{tabular}
  \label{tab:oracle-F}
\end{table}

It is also interesting to see the performance gap between our method and an oracle (having access to the optimal per-dataset threshold).
To this end we compare the \emph{average} performance of SIMFIT and \ours~to the \emph{maximum} performance fixed threshold (over the discrete set of thresholds $\{0.5, 0.75, 1, 1.5, 2, 3, 4\}$) for each dataset.
For MegaDepth1500 and ScanNet1500 we compute this optimal threshold per-model.
For IMC-PT we compute it per-scene (all scenes use SIFT).
We report results for Essential matrix estimation in \Cref{tab:oracle-E} and for Fundamental matrix estimation in~\Cref{tab:oracle-F}.
We find a drop of about 0.5 percentage points for our methods on average, while about a 1.5 point drop from SIMFIT.
In general, this drop is rather small compared to an incorrectly specified fixed threshold.

%% file: tables/E-poselib.tex
\begin{table*}
  \centering
  \caption{\textbf{Essential matrix estimation performance as a function of the threshold $\tau$ using \poselib.}}
  \begin{tabular}{lccccccc}
    \toprule
      & $\tau=0.5$ & $\tau=0.75$ & $\tau=1.0$ & $\tau=1.5$ & $\tau=2$ & $\tau=3.0$ & $\tau=4.0$ \\
    \midrule
    Fixed & 46.2 & 48.2 & 49.1 & 49.6 & 49.2 & 47.9 & 46.3 \\
    SIMFIT & 48.2 & 48.4 & 48.4 & 48.5 & 48.5 & 48.5 & 48.5 \\
    \ours~(Ours)& 49.2 & 49.5 & 49.6 & 49.7 & 49.6 & 49.6 & 49.6 \\
    \ours multi~(Ours)& 49.5 & 49.5 & 49.5 & 49.5 & 49.5 & 49.5 & 49.5 \\
    \bottomrule
  \end{tabular}
  \label{tab:E-poselib}
\end{table*}

%% file: tables/F-poselib.tex
\begin{table*}
  \centering
  \caption{\textbf{Fundamental matrix estimation performance as a function of the threshold $\tau$ using \poselib.}}
  \begin{tabular}{lccccccc}
    \toprule
      & $\tau=0.5$ & $\tau=0.75$ & $\tau=1.0$ & $\tau=1.5$ & $\tau=2$ & $\tau=3.0$ & $\tau=4.0$ \\
    \midrule
    Fixed & 31.3 & 33.2 & 34.0 & 33.8 & 33.1 & 31.3 & 29.7 \\
    SIMFIT & 33.2 & 33.8 & 34.0 & 34.2 & 34.3 & 34.3 & 34.3 \\
    \ours (Ours) & 34.5 & 34.6 & 34.7 & 34.7 & 34.8 & 34.8 & 34.7 \\
    \ours~multi (Ours) & 34.7 & 34.8 & 34.7 & 34.8 & 34.8 & 34.8 & 34.8 \\
    \bottomrule
  \end{tabular}
  \label{tab:F-poselib}
\end{table*}

%% file: sec/conclusion.tex
\section{Conclusion}
\label{sec:conclusion}
We have presented \ours, a method that significantly increases RANSAC's threshold robustness by estimating the inlier noise scale. 
We hope that our promising results will inspire additional research in how to estimate optimal thresholds.

\textbf{Limitations:}
\begin{enumerate}[label=\textbf{\alph*)}]
    \item The main assumption of our approach (and SIMFIT's), is that the optimal threshold is linearly related to the inlier noise scale by a factor ($F^{-1}(\alpha)$) independent of the data. Whether this holds in practice is not entirely clear. When comparing ours to an oracle, i.e., knowing the optimal threshold for a scene, ours typically has about 0.5\% to 1\% lower performance.
    \item While we identify two sources of bias which we remedy, we do not prove our estimator is unbiased. In fact, it is likely to be biased. As analytically proving estimators are unbiased even for very simple problems is difficult it is beyond the scope of the current work.
    \item Depending on the application, the increased runtime of our algorithm could be a potential issue. However, this could be remedied by only running the estimation on a small fraction of images, and reusing the estimate.
    \item 
    We use the matching data as provided by \posebench~in our experiments, which includes mostly phototourism and indoor type data. It is possible that for other types of data our method would perform worse.

\end{enumerate}